\documentclass[accepted]{article}
\pdfoutput=1

\usepackage{times}
\usepackage{graphicx} 
\usepackage{subfig} 
\usepackage{amsmath}
\usepackage{amssymb}
\usepackage{epstopdf}
\usepackage{natbib}
\usepackage{algorithm}
\usepackage{algorithmic}
\usepackage{hyperref}

\usepackage{icml2016}

\icmltitlerunning{Active Learning for Approximation of Expensive Functions with Normal Distributed Output Uncertainty}

\def\bfp{\mbox{\boldmath$p$}}
\def\bfg{\mbox{\boldmath$g$}}

\begin{document} 

\twocolumn[
\icmltitle{Active Learning for Approximation of Expensive Functions \\ with Normal Distributed Output Uncertainty}

\icmlauthor{Joachim van der Herten}{joachim.vanderherten@intec.ugent.be}
\icmlauthor{Ivo Couckuyt}{ivo.couckuyt@intec.ugent.be}
\icmlauthor{Dirk Deschrijver}{dirk.deschrijver@intec.ugent.be}
\icmlauthor{Tom Dhaene}{tom.dhaene@intec.ugent.be}
\icmladdress{Ghent University, Technologiepark 15, B-9052 Ghent, BELGIUM}

\icmlkeywords{FLOLA-Voronoi, Active Learning, Surrogate Modeling, Output Uncertainty}

\vskip 0.3in
]

\begin{abstract} 
When approximating a black-box function, sampling with active learning focussing on regions with non-linear responses tends to improve accuracy. We present the FLOLA-Voronoi method introduced previously for deterministic responses, and theoretically derive the impact of output uncertainty. The algorithm automatically puts more emphasis on exploration to provide more information to the models.
\end{abstract} 

\section{Introduction}
Consider an unknown multivariate black-box function $f: \mathbb{R}^d \rightarrow \mathbb{R}$. Our aim is obtaining an approximation of the response $\tilde{f}$ using machine learning methodology (regression) by evaluating $f$ on distinct points $\bfp_i \in \mathbb{R}^d$. The corresponding  responses are subject to normal distributed uncertainty $\lambda$: $y_i \sim \mathcal{N}(f(\bfp_i), \lambda)$ . However, assuming evaluation of $f$ involves a significant cost, only a limited amount of $N$ evaluations can be performed. Intelligent choice of these evaluations (\textit{experimental design}) can result in significantly better approximations.

Traditional Design of Experiments for computer experiments select all $N$ data points prior to  evaluation. These space-filling methodologies do not allow exploiting information obtained by each evaluation. In contrast, we present an active learning approach developed in the context of surrogate modeling known as FLOLA-Voronoi \cite{Vanderherten2015}. Starting from a small set of initial evaluations, iteratively new evaluation(s) are chosen based on an analysis of the previous responses. As a result, the algorithm increases the information density in non-linear areas that are more difficult to approximate. Originally the algorithm was formulated for deterministic responses, we analyze the impact of uncertain outputs theoretically in Section~\ref{sec:analysis}, and illustrate the effect in Section~\ref{sec:illustration}.

\section{Method}
\label{sec:analysis}
We present the FLOLA-Voronoi method, an improved version of the LOLA-Voronoi method \cite{Crombecq2010a} which has been applied successfully previously \cite{Deschrijver2011, Rosenbaum2012,Deschrijver2012}. The method can operate without intermediate model training, balances exploration and exploitation and remains efficient for high-dimensional functions. Two scores (computed by two algorithms) are aggregated for each previously evaluated point $\bfp$. The aggregated scores are ranked, and the highest ranked points are chosen to select a new point for evaluation local to the chosen point, using for instance the maximin criterion.

\begin{figure*}[t]
\centering
\subfloat[$\lambda=0$]{\includegraphics[width=0.38\textwidth]{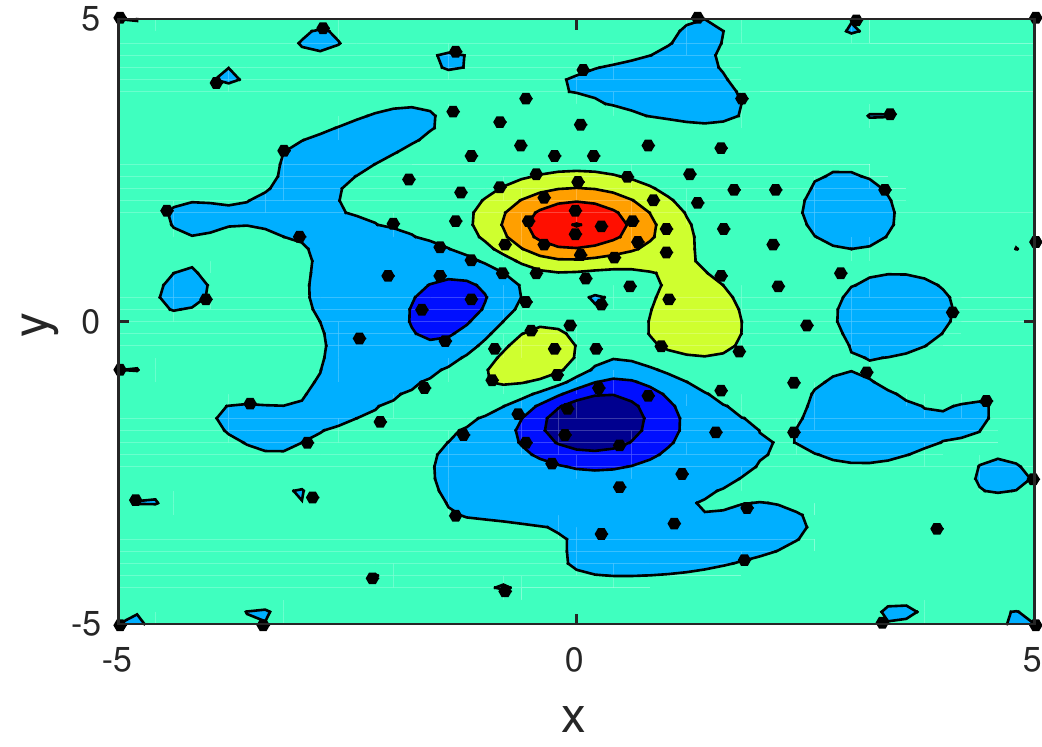}}%
~
\subfloat[$\lambda=1$]{\includegraphics[width=0.38\textwidth]{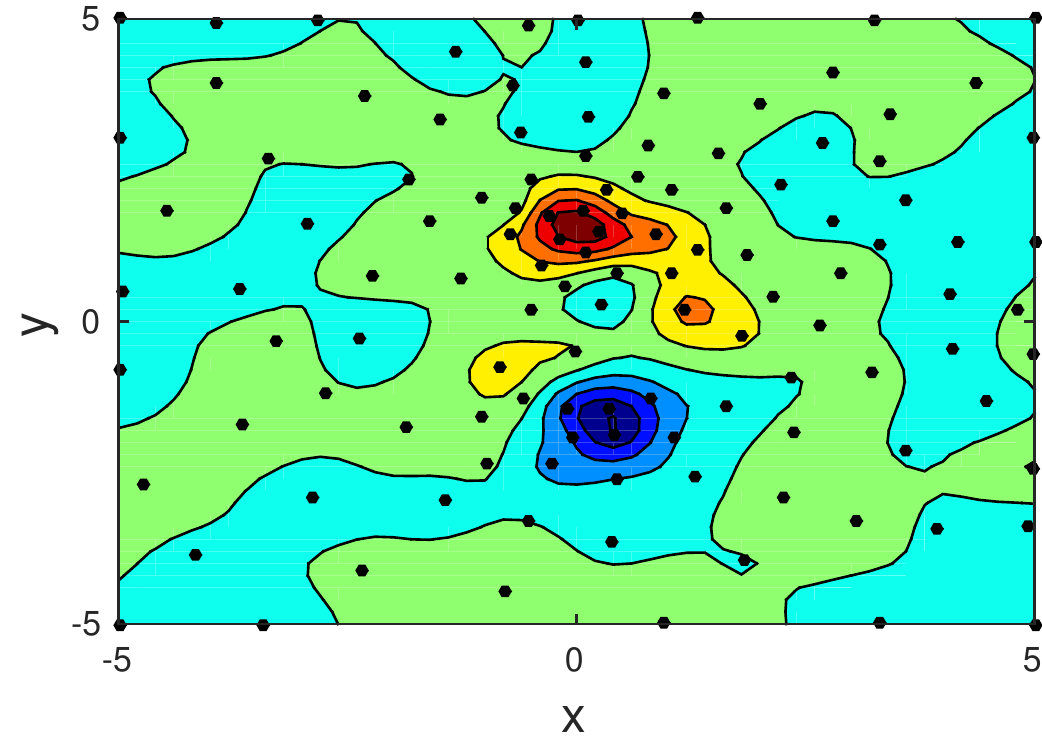}}%
\caption{FLOLA-Voronoi applied to the \textit{Peaks} problem, with and without noise. Clearly, the noiseless case is more focussed on the non-linear area whereas the noisy case results in more exploration.}
\label{fig:flola}
\end{figure*}

\subsection{Exploration}
The Voronoi exploration algorithm \cite{Crombecq2009a} estimates the size of the voronoi cells whose centroids corresponds to earlier evaluated data points: each point is assigned a score proportional to the relative size of the cell with respect to the other cells. As only a ranking of cell sizes is required, a Monte-Carlo estimate is sufficiently accurate and no full Voronoi tesselation needs to be computed. This algorithm only considers the d-dimensional input space and is insensitive to $\lambda$.

\subsection{Exploitation}
For a previously evaluated point $\bfp_r$, neighbouring points are selected to compute a gradient estimate $\bfg$ by solving a linear system of equations. Denote the set $N(\bfp_r)$ as the set containing the indices of the neighbouring points of $\bfp_r$. The original crosspolytope selection criterion presented in \cite{Crombecq2010a} does not scale well to high-dimensional input spaces, however this can be resolved by using a faster selection scheme \cite{Vanderherten2015}. The non-linearity error $E$ sums the differences between the linear prediction and the response obtained from the simulator.
This score depends on the output and therefore is sensitive to $\lambda$. We now show how much the uncertainty impacts $E$, under the assumption of normal distributed noise: $y_i =  f(\bfp_i) + \epsilon_i$ with $\epsilon_i \sim \mathcal{N}(0, \lambda)$. By expanding the terms and applying the triangular inequality:
\begin{eqnarray}
E(\bfp_r) & \leq & \sum\limits_{i \in N(\bfp_r)}^{} |f(\bfp_i) - (f(\bfp_r)+ \bfg . (\bfp_i-\bfp_{r}))|  \nonumber \\
\label{eq:gradienterrornoise}
& & + \sum\limits_{i \in N(\bfp_r)}^{} \zeta_i.
\end{eqnarray}
The upper bound of the score corresponds to the score for deterministic responses, complemented with a sum of noise terms $\zeta_i = |\epsilon_i - \epsilon_r|$. We denote the sum of the noise terms as random variable $X$. All $\epsilon$ are normally distributed with zero mean, so the difference of both variance terms is also normally distributed with zero mean and variance $2 \lambda$. This means each $\zeta_i$ is distributed according to a folded normal distribution:
\begin{equation*}
\zeta_i \sim \mathcal{FN} \left( 2 \sqrt{\frac{\lambda}{\pi}}, 2\lambda \left( 1-\frac{2}{\pi}\right) \right). 
\end{equation*}
The distribution of $X$ depends on $T = |N(\bfp_r)|$, and the shape of its probability density function resembles the log-normal distribution, its explicit formulation can be calculated but is a lengthy expression. Defining $u(t) = (t-1)^2+1$ and $v(t) = t^2-2t+2$, the expectancy and variance are given by following expressions:
\begin{eqnarray*}
\mathbb{E}\left[X\right] & = & \frac{2}{\pi}\sqrt{\lambda (\pi-2)}(T + (T-1) \sqrt{u(T)}), \\
\text{Var}\left[X\right] & = & \frac{2 \lambda (\pi - 2) \left( 4(T-1)+5\pi v(T)\right)}{\pi^2} - \mathbb{E}\left[X\right].
\end{eqnarray*}
The variance of $X$ perturbs the non-linearity score most. From the equation it can be derived that higher $\lambda$ and a higher $T$ cause more variance on the distribution of $X$. Using the neighbourhood selection mechanism of FLOLA-Voronoi the latter can be countered by including only the most significant neighbours in Equation~\ref{eq:gradienterrornoise} (instead of all).

\section{Illustration}
\label{sec:illustration}
The perturbations caused by $\text{Var}\left[X\right]$ impact the ranking of FLOLA-Voronoi causing more exploration, as the exploitation score becomes less decisive. This is illustrated by selecting 120 data points from the same function (\textit{Peaks}) twice, once with output noise and once without for a constant $T$. The sample distributions together with a contour plot of a fitted Kriging model are shown in Figure~\ref{fig:flola}. Clearly, the perturbed non-linearity scores cause more exploration which is a desirable property.

\section{Conclusion}
We presented the FLOLA-Voronoi algorithm for efficient model-independent sequential sampling of expensive black-box functions, and analysed the impact of uncertain response values on the algorithm. We have shown how the exploitation component of the algorithm is affected, and illustrated the resulting enhanced exploration. This is a desirable property in the presence of output uncertainty, as the model typically requires more information for linear areas.

\section*{Acknowledgements} 
This research has (partially) been funded by the Inter university Attraction Poles Programme BESTCOM initiated by the Belgian Science Policy Office. Ivo Couckuyt is a post-doctoral research fellow of the Research Foundation Flanders (FWO).

\bibliography{bibliography}
\bibliographystyle{icml2016}

\end{document}